\documentclass{article} 
\usepackage{iclr2026_conference,times}

\usepackage{amsmath,amsfonts,bm}

\def\eqref#1{equation~\ref{#1}}

\def\1{\bm{1}}

\DeclareMathAlphabet{\mathsfit}{\encodingdefault}{\sfdefault}{m}{sl}
\SetMathAlphabet{\mathsfit}{bold}{\encodingdefault}{\sfdefault}{bx}{n}

\usepackage{amsmath}
\usepackage{amssymb}
\usepackage{booktabs}
\usepackage[most]{tcolorbox}
\newtcolorbox{promptbox}[1]{
  enhanced,
  colback=gray!5,
  colframe=blue!60,
  boxrule=0.8pt,
  arc=4pt,
  left=6pt,
  right=6pt,
  top=6pt,
  bottom=6pt,
  fonttitle=\bfseries,
  title={#1},
  breakable
}

\usepackage{color}
\usepackage{multirow}

\usepackage{url}
\usepackage{hyperref}

\title{PersRM-R1: Enhance Personalized Reward Modeling with Reinforcement Learning}

\author{%
\hspace{4em} 
Mengdi Li\textsuperscript{1,2,*} 
\hspace{4em}
Guanqiao Chen\textsuperscript{3,*}  
\hspace{5em}
Xufeng Zhao\textsuperscript{4} 
\hspace{8em}
\And
\hspace{4em}
Haochen Wen\textsuperscript{5}  
\hspace{5em}
Shu Yang\textsuperscript{1,2}  
\hspace{6.5em}
Di Wang\textsuperscript{1,2,†} 
\hspace{5em} 
\\
[1.5ex]
\parbox{\textwidth}{%
\footnotesize
\textsuperscript{1}Provable Responsible AI and Data Analytics Lab, \textsuperscript{2}King Abdullah University of Science and Technology \\[0.3ex]
\textsuperscript{3}University of Science and Technology of China, \textsuperscript{4}University of Hamburg, \textsuperscript{5}University College London \\[0.5ex]
*~Equal Contribution \quad †~Corresponding Author
}
}

\iclrfinalcopy 
\begin{document}

\maketitle

\begin{abstract}
Reward models (RMs), which are central to existing post-training methods, aim to align LLM outputs with human values by providing feedback signals during fine-tuning. However, existing RMs struggle to capture nuanced, user-specific preferences, especially under limited data and across diverse domains. Thus, we introduce PersRM-R1, the first reasoning-based reward modeling framework specifically designed to identify and represent personal factors from only one or a few personal exemplars. To address challenges including limited data availability and the requirement for robust generalization, our approach combines synthetic data generation with a two-stage training pipeline consisting of supervised fine-tuning followed by reinforcement fine-tuning. Experimental results demonstrate that PersRM-R1 outperforms existing models of similar size and matches the performance of much larger models in both accuracy and generalizability, paving the way for more effective personalized LLMs.\footnote{The code and datasets are available at \url{https://github.com/Jeffrey-Guanqiao/PersRM-R1}.}
\end{abstract}


\section{Introduction}

\begin{figure}[ht!]
\centering
\includegraphics[width=0.85\columnwidth]{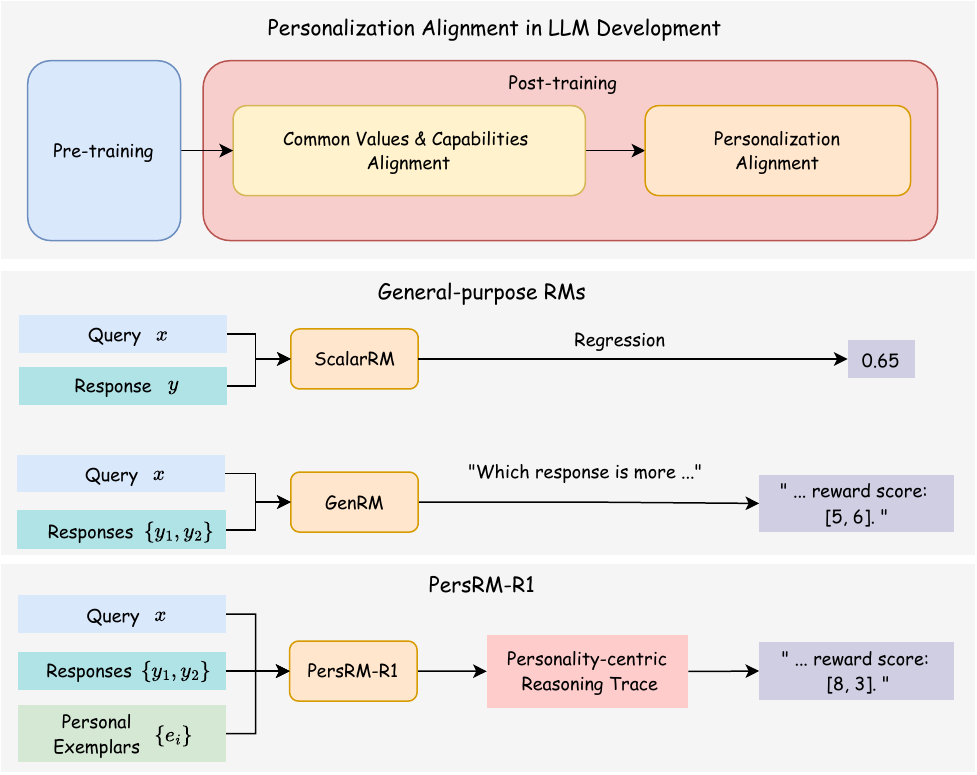}
\caption{
\textbf{Top}: 
Post-training for user-facing LLMs involves two key stages: 1) aligning the LLM with shared human values and endowing it with fundamental capabilities; 2) tailoring the LLM to individual preferences.
\textbf{Middle}: General-purposed RMs, including both scalar and generative types, are trained on standard tasks to generate reward scores for given responses.
\textbf{Bottom}: PersRM-R1 is capable of capturing subtle personality traits presented in personal exemplars, conducting personality-centric reasoning analyses, and generating reward scores that reflect how closely each response aligns with the style exhibited in the personal exemplars. 
}
\label{fig:llm-dev-pipeline-and-persrm}
\end{figure}

The paradigm of pre-training followed by post-training has been widely adopted in both academia and industry for developing large language models (LLMs). 
In the post-training stage, common values, such as harmlessness, helpfulness, and honesty, that are shared among human beings, and common capabilities, such as chatting, reasoning, which are shared among various tasks, have been the optimization objective in model fine-tuning \citep{bai2022TrainingHelpfulHarmless, bai2022ConstitutionalAIHarmlessness, ouyang2022TrainingLanguageModels, stiennon2020LearningSummarizeHuman}. 
However, as LLMs are increasingly integrated into personalized applications, e.g., personal assistants, tutoring systems, and writing aids, there is a growing need for models that can not only follow specific user instructions but also align closely with individual preferences and communication styles.
These differences, such as preference for conciseness, use of humorous expressions, and opinions on controversial topics, distinguish one person from another. 
Thus, \emph{personalization alignment} that makes the user-facing LLM best fit for each individual is essential; however, it has not received much attention in the community (cf.~Fig.~\ref{fig:llm-dev-pipeline-and-persrm}, top panel). 

Reinforcement learning (RL) methods based on reward models (RMs) are commonly applied in LLM post-training, where a reward model is trained to provide reward signals for the training of the policy model through reinforcement learning algorithms \citep{christiano2017DeepReinforcementLearning, ouyang2022TrainingLanguageModels}.
The reliability of RMs lies at the core of the success of such methods \citep{li2023InternallyRewardedReinforcement, casper2023OpenProblemsFundamental}. 
RMs for common values and capabilities alignment are trained on large-scale crowd-sourcing data to represent homogeneous preferences for general purposes (cf.~Fig.~\ref{fig:llm-dev-pipeline-and-persrm}, middle panel). In contrast, RMs for individuals ought to be personalized, as studied in the literature of neuroscience, where the emission of rewarding signals, like dopamine, is influenced by the value component that reveals the brain’s \emph{subjective evaluation} of the effects of goal achievements \citep{schultz2015NeuronalRewardDecision}. 

Developing such models, however, is non-trivial, particularly in settings where only limited user-specific data is available. In this work, we investigate how to construct and train personalized reward models under realistic data constraints. We focus on two central technical challenges: (a) the \emph{scarcity of individual-specific demonstrations} for effective model tuning, and (b) the base model’s \emph{insufficient sensitivity to nuanced personality traits} reflected in both exemplars and generated responses.

\begin{figure*}[ht]
\centering
\includegraphics[width=1.0 \textwidth]{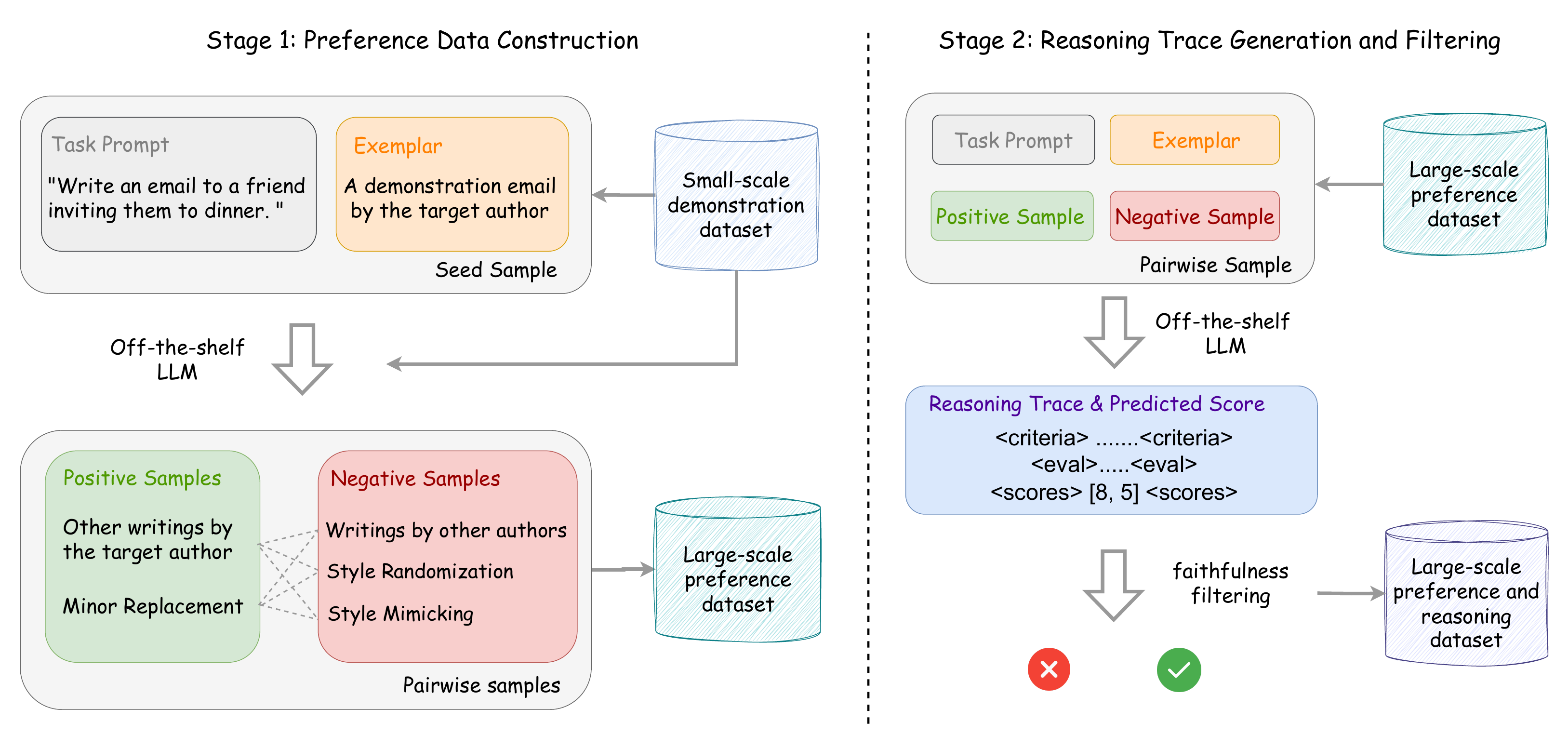}
\caption{Pairwise preference data generation pipeline includes: \textsc{Left}) personalization data argumentation with contrastive prompting (Stage 1), and \textsc{Right}) reasoning traces collection and post-hoc data filtering (Stage 2). }
\label{fig:persrm-r1-data generation pipeline}
\end{figure*}

We introduce a novel approach to address these challenges, ultimately resulting in our RMs, namely PersRM-R1. The overall development pipeline of our method is illustrated in Fig.~\ref{fig:training-pipeline}. 
Specifically, to tackle the first challenge, we design a synthetic data generation pipeline to enhance pairwise preference data (cf.~Fig.~\ref{fig:persrm-r1-data generation pipeline}, left panel).
For the second challenge, we propose to incorporate a \emph{personality-centric reasoning process} into reward modeling. This is motivated by our hypothesis that personality traits are inherently embedded within the text and must be explicitly extracted and analyzed to more accurately assess personality similarity between a personal exemplar and an arbitrary response. 
To this end, we generate synthetic reasoning traces for pairwise preference data (cf.~Fig.~\ref{fig:persrm-r1-data generation pipeline}, right panel), which are used to tune the base model in supervised fine-tuning (SFT) to enhance its foundational capability to reason about personality traits and to produce reward scores in a standardized format. 
Subsequently, reinforcement fine-tuning (RFT) is employed to further enhance its performance and generalizability, building on recent advances in reasoning-augmented LLMs \citep{deepseek-ai2025DeepSeekR1IncentivizingReasoning, liu2025InferenceTimeScalingGeneralist}. 

The performance of PersRM-R1 is evaluated on the representative personalization alignment task of personal stylish writing, across diverse genres including email, essays, news articles, blogs, and more. 
Experimental results show that PersRM-R1 only only outperforms existing models of comparable size but also matches the accuracy and generalizability of much larger models.
To the best of our knowledge, this work presents the first reasoning-based reward model tailored for personalization. We propose PersRM-R1, which integrates guided data augmentation with a two-stage fine-tuning pipeline to enable preference modeling from minimal user input. Beyond establishing a new direction for personalized alignment, our study systematically investigates alternative approaches such as in-context learning and conducts ablations to isolate the contributions of each pipeline component, providing a comprehensive foundation for future work in personalized reward modeling.

                                                                                                
\section{Related Work}

\noindent\textbf{LLM Personalization.} Existing methods for personalizing LLMs can be categorized into two classes:
1) tuning-free methods, such as retrieval-augmented generation (RAG) \citep{salemi2024LaMPWhenLargea}, prompt engineering \citep{park2023GenerativeAgentsInteractive}, and steering vector intervention \citep{konen2024StyleVectorsSteering, cao2024PersonalizedSteeringLargea}; 
2) tuning-based methods, which influence the manifested personality traits of LLMs through supervised fine-tuning \citep{shao2023CharacterLLMTrainableAgent}, direct preference optimization \citep{li2024PersonalizedLanguageModeling, zeng2024PersLLMPersonifiedTraining, shaikh2025ShowDontTell}, and RM-based approaches \citep{chen2025PALSampleEfficientPersonalized}. 
Tuning-free methods, while simple to implement, often induce extra token costs and sacrifice inference speed. Moreover, their performance can be inconsistent and sensitive to retrieved information, prompts, or steering vectors. 
In contrast, tuning-based methods that directly integrate personality traits into the model parameters offer a more robust and fundamental solution. 
A concurrent work by \citet{chen2025PALSampleEfficientPersonalized} focuses on developing personalized RMs by efficiently tuning parameters of the base model for each specific traits, a strategy that becomes impractical when considering the vast number of potential users. In contrast, our work focuses on a distinct path to personalized reward modeling, where we develop a unified RM that captures personal traits from individual exemplars and arbitrary responses, producing reward scores that reflect personality similarity.

\noindent\textbf{Reasoning-enhanced Reward Modeling.} 
RMs in current literature can be categorized into two classes: scalar and generative RMs \citep{lambert2024rewardbenchevaluatingrewardmodels, liu2025InferenceTimeScalingGeneralist}. 
Scalar RMs are trained to predict a scalar reward value given a pair of prompt and response. 
In contrast, generative RMs offer a more general formulation for reward modeling that exhibits strong generalization potential. In addition, reasoning capability can be naturally incorporated into the rewarding process \citep{yang2024ReinforcingThinkingReasoningEnhanced, chen2025RMR1RewardModeling, liu2025InferenceTimeScalingGeneralist, guo2025RewardReasoningModel}, which has attracted increasing attention recently, motivated by the success of incorporating long-reasoning process in LLMs in tasks of coding and math problem solving \citep{deepseek-ai2025DeepSeekR1IncentivizingReasoning, openai2024OpenAIO1System,Zhao24EnhancingZeroshot}. 
Most of existing work in this direction focus on tasks where ground-truth rule-based rewards can be obtained. 
In contrast, employing reinforcement learning in open domains is more challenging as reliable rewards are absent. 
While previous work \citep{liu2025InferenceTimeScalingGeneralist} employs self-generated principles to generate rewards aimed at common value alignment, we demonstrate that, with appropriate modifications to address challenges such as identifying nuanced preference characteristics and coping with limited data, similar strategies can be adapted for personalized reward generation.

\section{Methodology}

\begin{figure}[t]
\centering
\includegraphics[width=0.65\columnwidth]{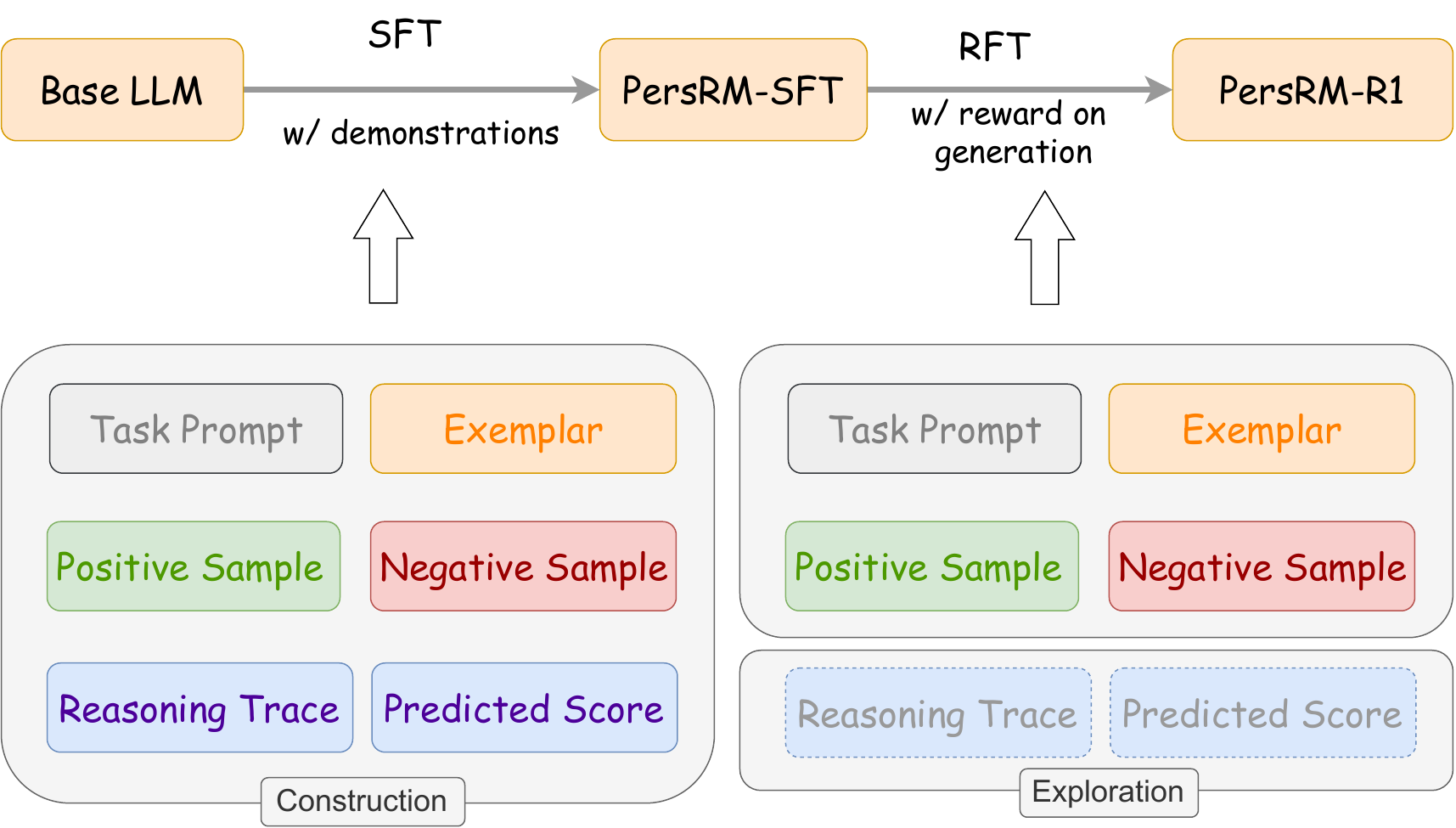}
\caption{
    Training pipeline of PersRM-R1. The process begins with personalization data augmentation (see Sec.~\ref{subsec:data-curation}), followed by supervised fine-tuning (Sec.~\ref{subsec:sft}), and concludes with reinforcement fine-tuning (Sec.~\ref{subsec:rft}) to explore effective reasoning traces.
}
\label{fig:training-pipeline}
\end{figure}

In this section, we introduce the three key phases in developing personalization reward models: data curation (cf. Sec.~\ref{subsec:data-curation}), supervised fine-tuning (SFT, cf. Sec.~\ref{subsec:sft}), and reinforcement fine-tuning (RFT, cf. Sec.~\ref{subsec:rft}).
Each subsection herein addresses distinct challenges introduced by RM in the context of personalization alignment, while preliminaries on SFT and RFT are provided in Appendix~\ref{app:preliminary}.
An overview of the full training pipeline is shown in Figure~\ref{fig:training-pipeline}.

\subsection{Personalization Data Augmentation}
\label{subsec:data-curation}

To address the challenge of lacking user-specific preference data for reward modeling, 
we employ data augmentation techniques using LLMs to generate a synthetic dataset conditioned on a limited user corpus \( \mathcal{D}_\text{expl} = \{ e_i \} \).
First, LLMs are prompted to produce both aligned and divergent responses conditioned on user exemplars. Second, deeper reasoning is elicited to re-evaluate the auto-labeled responses generated in the first stage, enabling more faithful scoring. Together, these two stages distill the LLMs' understanding of user-specific preferences, producing high-quality synthetic data sufficient for fine-tuning the reward model.

\noindent\textbf{Pairwise Preference Data Construction.} Given a query/problem $x$, we prompt an LLM to generate content that is close ($y^+$), or divergent ($y^-$) to the user's personality traits exhibited in the exemplars/context $e$, respectively, resulting in a synthetic collection  \(\mathcal{D}_\text{syn} = \{ x, e, y^+, y^- \} \).
\begin{itemize}
    \item  To generate positive samples \( y^+ \) that preserve user-specific stylistic characteristics, we employ two strategies. The first, \emph{intra-author retrieval}, involves selecting alternative responses authored by the same user for different queries, thereby ensuring authentic and consistent personalized style. The second, \emph{lexical perturbation}, creates controlled variants of demonstrations through (up to six) synonym substitutions while maintaining the original grammatical structure and sentence order.

    \item Negative samples \( y^- \) are obtained via three complementary approaches. \emph{Cross-author retrieval} sources responses from different users, introducing clear stylistic and preference divergences. \emph{Random sampling} employs large language models to generate loosely related or off-topic responses, serving as weak negative examples.
    Finally, \emph{confounding sampling} leverages LLMs to generate responses that approximate the stylistic features present in the user’s exemplars. While the current LLMs lack the ability to faithfully capture personalized styles, a gap this work seeks to address, these samples nonetheless serve as strong adversarial negatives, encouraging the reward model to attend to fine-grained stylistic cues.
    
\end{itemize}

This comprehensive sampling framework spans a continuum of difficulty levels, which is essential for developing robust and generalizable reward models~\citep{li2025CurriculumRLAIFCurriculumAlignment}.

\noindent\textbf{Reasoning Trace Generation.} Reward models augmented with explicit reasoning traces have demonstrated superior capabilities in delivering fine-grained and reliable preference assessments~\citep{li2025CurriculumRLAIFCurriculumAlignment, lee2024RLAIFVsRLHF}.
To further enhance alignment with user-specific preferences, we elicit comparative reasoning from LLMs to assess the stylistic match or mismatch between generated responses.

Specifically, given a quadruple \( (x, e, y^+, y^-) \) from the synthetic dataset \( \mathcal{D}_\text{syn} \), we prompt the LLM to evaluate the similarity between each response (\( y^{+} \) and \( y^{-} \)) and the user exemplars \( e \) respectively.
The model is constrained to output a reasoning-based evaluation tuple \( \mathcal{V} = (\tau, r^+, r^-) \), where:
\begin{itemize}
    \item \( \tau \) is a step-by-step reasoning trace explaining the comparative judgment, including an analysis of stylistic alignment, tone, phrasing, or semantic intent;
    \item \( r^+ \) and \( r^- \) are scalar reward scores, typically normalized within a bounded range (e.g., [1, 10], reflecting the model’s confidence in the stylistic compatibility of \( y^+ \) and \( y^- \) with the user exemplars.
\end{itemize}

This encourages the model to articulate fine-grained distinctions and builds interpretability into reward modeling by exposing the rationale behind preference decisions. 

\noindent\textbf{Faithful Reasoning Trace Filtering.} To ensure alignment between reasoning traces and the intended preference signal, we apply a filtering step that retains only \textit{faithful} reasoning outputs, i.e. those that are consistent with the preference implied by the contrastive generation stage (i.e., \( y^{+} \) should be stylistically more aligned with the exemplars \( e \) than \( y^{-} \)). Reasoning traces that contradict this assumption are excluded from the dataset, as they may introduce conflicting supervision and hinder effective training. 
This filtering ensures coherence between the scalar reward scores and the reasoning trace \( \tau \), resulting in the dataset \( \mathcal{D}_\text{SFT} = \{ x, e, y^{+}, y^{-}, \mathcal{V} \} \), which contains contrastive response pairs with faithful justifications. This dataset is used for subsequent supervised fine-tuning (SFT) to elicit reasoning justifications that improve the reward model’s ability to assess preferences with increased accuracy and interpretability.

\subsection{Supervised Fine-Tuning}  
\label{subsec:sft}

With the curated dataset \( \mathcal{D}_\text{SFT} \) in place, we first perform supervised fine-tuning on the RM to internalize the knowledge extracted through guided prompting. This results in a warm-started RM, denoted as PersRM-SFT, that can better assess user-aligned preferences in a generative way. The training objective is to maximize the conditional log-likelihood
\[
\max_{\theta} \; \mathbb{E}_{(x, e, y^+, y^-, \mathcal{V}) \sim \mathcal{D}_\text{SFT}} \; \log p_{\theta}(\mathcal{V} \mid x, y^+, y^-, e),
\]
where \( \mathcal{V} \) is the structured reasoning-augmented evaluations and \( \theta \) denotes the trainable parameters of the reward model. This SFT phase serves to align the model’s outputs with high-quality, faithful supervision signals, preparing it for subsequent reinforcement learning. 

\subsection{Reinforcement Fine-Tuning}
\label{subsec:rft}

Following supervised fine-tuning, we further optimize the reward model through RL to promote exploration of diverse reasoning patterns and improve robustness in preference assessments. While SFT provides high-quality supervision signals, it remains limited to the imitation of static patterns seen during data curation. In contrast, RFT enables the model to generate novel reasoning traces that generalize beyond the curated examples, allowing for more adaptive and discriminative preference modeling.

\noindent\textbf{Sampling and Format Validation.}  Given an input quadruple \( (x, e, y^+, y^-) \), we sample reasoning-based evaluations \( \mathcal{V} \sim p_\theta(\mathcal{V} \mid x, y^+, y^-, e) \) from the current reward model. To ensure valid learning signals, we first perform format validation using a deterministic procedure \( \texttt{fmt}(\mathcal{V}) \), which checks whether the generated output conforms to the expected structured format. If the format is valid, we parse the reasoning trace and scores as \( (\hat{\tau}, \hat{r}^+, \hat{r}^-) \leftarrow \mathcal{V} \); otherwise, the response is considered invalid and penalized.

\noindent\textbf{Reward Function Design.} To guide the model toward both faithful reasoning and correct preference assessments, we define a sparse reward function based on \emph{format correctness} and \emph{consistency with contrastive supervision}:
\begin{equation} \label{eq:reward_function}
r(\mathcal{V}, x, y^+, y^-, e) = 
\begin{cases} 
  -1, & \neg \texttt{fmt}(\mathcal{V}) \\
   0, & \texttt{fmt}(\mathcal{V}) \wedge \hat{r}^+ \leq \hat{r}^- \\
   +1, & \texttt{fmt}(\mathcal{V}) \wedge \hat{r}^+ > \hat{r}^-. \\
\end{cases}
\end{equation}
This reward signal encourages the generation of well-structured outputs that faithfully reflect the intended preference signal, where the preferred response \( y^+ \) should receive a higher score than \( y^- \). Outputs that violate formatting constraints or misalign with the contrastive signal receive a penalty, thereby discouraging hallucinated or unreliable evaluations.

\noindent\textbf{Policy Optimization.} The reward model is then updated using standard RL algorithms with the objective of maximizing expected reward:
\[
\max_{\theta} \; \mathbb{E}_{(x, e, y^+, y^-) \sim \mathcal{D}_\text{syn}, \mathcal{V} \sim p_\theta} \; r(\mathcal{V}, x, y^+, y^-, e).
\]
This process results in a reinforcement-tuned personalized reward model, denoted as PersRM-R1, which benefits from both SFT pretraining and RL-guided exploration. To assess the necessity of SFT, we include an ablation variant, denoted as PersRM-RFT, which applies RL directly on the pretrained base model without SFT initialization. This cold-start configuration allows us to evaluate whether reinforcement alone suffices to capture user-specific preferences or if the curated supervision from SFT is critical for performance.


\section{Experiments}

\subsection{Experimental Setup}

\noindent\textbf{Dataset.}
Our experiments use two datasets centered on writing styles: the \textbf{CCAT} dataset, which contains news articles by 50 authors \citep{lewis2004RCV1NewBenchmark}, and \textbf{CMCC} dataset, which includes multi-genre writings such as emails, blogs, essays by 21 authors \citep{goldstein-stewart2008CreatingUsingCorrelated}. 
Our evaluation protocol is designed to test generalization to both unseen authors and unseen genres, necessitating a strict, non-overlapping partition of data into training, validation, and test sets.
We employ a \emph{strictly author-disjoint split}, ensuring authors used for training, validation, and testing are entirely separate. This allows our evaluation to assess the model's true capabilities in personality trait analysis and scoring, rather than relying on memorization. 

Additionally, we develop a challenging cross-domain test set to assess the generalizability of models in measuring personality similarity across different genres. 
An ideal model should perform well in this scenario, as we hypothesize that personal traits remain consistent cross domains. 
This test set is constructed exclusively from CMCC, which features multiple genres for the same author, whereas CCAT contains only news articles. 

\begin{itemize}
\item \textbf{Training Set}: 
We construct the training set using our data curation pipeline based on news articles, emails, and essays of a large group of authors, comprising \emph{45 authors from the CCAT corpus} and \emph{18 authors from the CMCC corpus}, which results in approximately 17.2k pairwise preference and reasoning samples. 
\item \textbf{Validation Set}: The validation set is built using a held-out group of authors: \emph{2 authors from CCAT} and \emph{1 author from CMCC}. This set consists of 200 samples.
\item \textbf{Standard Test Set}: The test set is composed of a separate group of held-out authors: the \emph{remaining 3 authors from CCAT} and the \emph{remaining 2 authors from CMCC}. This set contains 334 pairwise samples.
\item \textbf{Cross-Domain (Cr. Do.) Test Set}: This test set is constructed using documents from the \emph{3 CMCC authors} from the validation and test splits. Crucially, we exclusively use genres that are \emph{entirely withheld} from both the training and validation sets. These unseen genres include \emph{blog articles, interview transcripts, and chat logs}. This challenging set consists of 439 pairwise samples.
\end{itemize}

\noindent\textbf{Models.} 
We use Qwen2.5-3B-Instruct and Qwen2.5-7B-Instrcut as base models, as prior literature demonstrates that the strong fundamental reasoning abilities of Qwen series facilitate effective post-training and adaptation to reasoning-focused downstream tasks \cite{xie2025LogicRLUnleashingLLM}. 
Our resulting models are referred to as PersRM-R1-3B and PersRM-R1-7B, respectively. 
Qwen2.5-72B-Instruct~\citep{qwen2025Qwen25TechnicalReport} is employed for both preference data construction and reasoning trace generation.\footnote{DeepSeek-R1-Distill-Llama-70B \citep{deepseek-ai2025DeepSeekR1IncentivizingReasoning}, another strong open-source LLM known for its reasoning capabilities, is also examined. However, in our experimental scenario, its generated reasoning traces often lack proper formatting. In contrast, Qwen2.5-72B-Instruct exhibits satisfactory performance in both reasoning quality and adherence to formatting requirements.}

\noindent\textbf{Model Training.}
Our models are trained following the SFT and RFT procedure introduced in Sec.~\ref{subsec:sft} and Sec.~\ref{subsec:sft}. Further training details, such as hyperparameter setups, are provided in Appendix~\ref{app:training_and_testing}. 

\noindent\textbf{Evaluation Metric.}
Following previous work \citep{stiennon2020LearningSummarizeHuman, lambert2024rewardbenchevaluatingrewardmodels}, we evaluate RM performance using \emph{accuracy} with respect to the ground-truth preference labels, where random guessing yields an accuracy of 50\%. 
Further details on the evaluation methodology for different types of RMs are provided in Appendix~\ref{app:training_and_testing}.

\subsection{Baselines}
We compare PersRM-R1 with a diverse set of well-known reward models across three baseline categories:
(1) \textbf{Scalar RMs}: 
Internlm2-7B-Reward, RM-Mistral-7B, Skywork-Reward-Llama3.1-8B (abbreviated as SR-Llama3.1-8B), all of which are top-performing models of approximately 7B parameters on the RewardBench leaderboard~\citep{lambert2024rewardbenchevaluatingrewardmodels};
(2) \textbf{Generative RMs}: 
Mistral-v0.3-7B-Instruct~\citep{jiang2023Mistral7B}, the Qwen2.5 series (ranging from 7B to 32B), Llama-3.1-8B-Instruct, Llama-3.1-70B-Instruct~\citep{grattafiori2024Llama3Herd}, representing mainstream open-source LLMs across a large range of model size;
(3) \textbf{Reasoning RMs}: 
We also compare against two recently released, state-of-the-art RMs that have been specifically enhanced for reasoning capabilities: RM-R1-7B~\citep{chen2025RMR1RewardModeling} and RRM-7B~\citep{guo2025RewardReasoningModel}.

\subsection{Experimental Results}
Table~\ref {tab:performance_comparison} presents a performance comparison between our models and various baselines. 
We report the performance on the standard test set, with separate results for each dataset (CCAT and CMCC), as well as on the cross-domain test set (Cr. Do.).

\begin{table}[ht]
\centering
\scalebox{0.9}{
\begin{tabular}{p{1.8cm}lccc}
\toprule
\textbf{Category} & \textbf{Model} & \textbf{CCAT} & \textbf{CMCC} & \textbf{Cr. Do.} \\
\midrule
\multirow{3}{=}{\centering Scalar RMs} 
    & Internlm2-7B-Reward & 67.8 & 69.2 & 64.3 \\
    & RM-Mistral-7B       & 65.7 & 68.1 & 62.8 \\
    & SR-Llama3.1-8B      & 65.3 & 68.4 & 68.8 \\
\midrule
\multirow{6}{=}{\centering Generative RMs} 
    & Mistral-v0.3-7B-Ins.  & 73.9 & 76.2 & 74.8 \\
    & Qwen2.5-7B-Ins.       & 77.6 & 75.8 & 72.4 \\
    & Qwen2.5-14B-Ins.      & 82.3 & 82.8 & 83.1 \\
    & Qwen2.5-32B-Ins.      & 86.4 & 87.1 & 86.7 \\
    & Llama3.1-8B-Ins.      & 77.3 & 79.3 & 73.8 \\
    & Llama3.1-70B-Ins.     & \textbf{94.3} & \underline{94.3} & \textbf{93.7} \\
\midrule
\multirow{4}{=}{\centering Reasoning RMs}
    & RM-R1-7B              & 89.8 & 88.2 & 89.7 \\
    & RRM-7B                & 87.2 & 89.6 & 89.3 \\
    & {PersRM-R1-3B (Ours)} & 91.8 & 92.2 & 89.7 \\
    & {PersRM-R1-7B (Ours)} & \underline{93.8} & \textbf{94.6} & \underline{92.3} \\
\bottomrule

\end{tabular}
}
\caption{
Performance comparison of various models across different test sets, reported as reward modeling accuracy (\%), where higher values indicate better performance. Best results are in bold, runner-up in \underline{underline}.
}
\label{tab:performance_comparison}
\end{table}

\noindent\textbf{Performance of Baselines.} 
Our evaluation of the baseline models yields three important findings. First, existing scalar-based RMs perform poorly in scoring personal preferences, with accracies below 70.0\%, indicating their inadequacy in distinguishing personalization divergence. 
Second, reward modeling performance for personalization exhibits a clear scaling trend, with larger models consistently achieving higher accuracy. For example, the Qwen2.5 series shows improved performance as model size increases, and Llama3.1-70B-Instruct achieves the best results across most benchmarks, underscoring the effectiveness of greater model capacity in capturing user-specific preferences.
Third, reasoning-based reward models consistently outperform non-reasoning generative models of comparable size. For example, RM-R1-7B achieves nearly 10\% higher accuracy than Qwen2.5-7B and Llama3.1-8B, and performs comparably to the much larger Qwen2.5-32B. 
These results underscore the superiority and potential of reasoning reward models in personalization alignment.

\noindent\textbf{Effectiveness of PersRM-R1.} 
Our proposed model, PersRM-R1-7B, achieves 93.8\% accuracy on CCAT and 94.6\% on CMCC, substantially outperforming other reasoning-based models of similar size. This demonstrates the effectiveness of our training paradigm, even with a limited amount of personalized data. Moreover, despite having significantly fewer parameters, our models approach the performance of much larger models such as Llama3.1-70B-Instruct, highlighting their efficiency and scalability.
These findings support our hypothesis that existing reward models, typically trained on datasets focused on mathematical or common-sense reasoning, lack the architectural and data-centric design necessary for capturing personalized preferences. This highlights the need for specialized personalization reward models, as advanced in this study.

\noindent\textbf{Cross-Domain Generalizability.}
The cross-domain test setup is challenging, especially for RMs with relatively small size. 
{PersRM-R1-7B} achieves an accuracy of 92.3\%, while the smaller {PersRM-R1-3B} attains 89.7\%, exhibiting strong cross-domain generalizability to genres not included in the training data. 
This strong performance on unseen genres highlights our method's ability to learn the underlying principles of personal preference rather than merely overfitting to the topics in the training data. This capability is crucial for building RMs that are practical and reliable in real-world applications.

\noindent\textbf{Generalizability to Additional Exemplars.}
We evaluate the generalizability to additional exemplars in inference time, as in realistic scenarios multiple exemplars could be available for a target user. 
We observe from experimental results in Table~\ref{tab:eval_number_exemplars} that incorporating additional personal exemplars leads to performance improvements across all models. This effect is particularly pronounced for Qwen2.5-7B-Instruct, which starts with relatively low performance when using only one exemplar. Notably, PersRM-R1-7B demonstrates even greater improvement, despite already achieving the best results in the single-exemplar setting, when compared to Llama3.1-70B-Instruct. 
This highlights the strong generalizability of PersRM-R1-7B with respect to the number of exemplars provided, even tuned on preference data with only one exemplar. 
Additional experimental results regarding different training paradigms and additional exemplars are provided in Appendix.~\ref{app:three-shot-generalizability-training-paradigms}.

\begin{table}[ht]
\centering
\footnotesize
\begin{tabular}{lclll}
\toprule
\textbf{Model} & \textbf{\# Ex.} & \textbf{CCAT} & \textbf{CMCC} & \textbf{Cr. Do.} \\
\midrule
\multirow{2}{*}{Qwen2.5-7B-Ins.} & 1 & 77.6 & 75.8 & 72.4 \\
 & 3 & 78.3 \scriptsize{(+0.7)} & 79.2 \scriptsize{(+3.4)} & 76.2 \scriptsize{(+3.8)} \\
\hline
\multirow{2}{*}{Llama3.1-70B-Ins.} & 1 & {94.3} & {94.3} & {93.7} \\
 & 3 & \textbf{94.3} \scriptsize{(+0.0)} & \underline{94.6} \scriptsize{(+0.3)} & \underline{93.9} \scriptsize{(+0.2)} \\
\hline
\multirow{2}{*}{PersRM-R1-7B} & 1 & {93.8} & {94.6} & 92.3 \\
 & 3 & \underline{94.1} \scriptsize{(+0.3)} & \textbf{95.1} \scriptsize{(+0.5)} & \textbf{92.6} \scriptsize{(+0.3)} \\
\bottomrule
\end{tabular}
\caption{
Reward modeling accuracy (\%) with varying numbers of exemplars (\# Ex.). Values in parentheses indicate improvement over the one-exemplar setting.
}
\label{tab:eval_number_exemplars}
\end{table}


\section{Analysis}
In this section, we present a series of analytical experiments to provide deeper insights into our proposed training methodology. The primary goal is to investigate the underlying reasons for its effectiveness and to characterize the dynamic changes that occur throughout the training process. 
\subsection{Effectiveness of Different Paradigms} \label{sec:training-paradigms}

We explore the effectiveness of different training paradigms: 1) SFT only, 2) RFT only, and 3) SFT followed by RFT (SFT+RFT).
Our experiments are conducted based on the {Qwen2.5-7B-Instruct} model and previously constructed datasets.
As illustrated in Table~\ref{tab:performance_ablation}, both standalone SFT and RFT are effective in enhancing the model's capabilities in measuring personality similarity. However, the paradigm of SFT followed by RFT, embodied in our {PersRM-R1-7B} model, delivers the most significant performance boost, substantially outperforming all other approaches. 
This highlights the value of integrating both supervised and reinforcement fine-tuning strategies, and suggests that the strong capabilities of PersRM-R1 are not solely due to knowledge distillation through SFT from the off-the-shelf LLM used in data construction, but also stem from the reinforcement learning procedure, which enables the model explore and generalize to novel skills for solving the task.

\begin{table}[ht]
\centering
\footnotesize
\begin{tabular}{cccccc}
\toprule
\textbf{Training Paradigm} & \textbf{CCAT} & \textbf{CMCC} & \textbf{Cr. Do.} \\ 
\midrule
Base Model & 77.6 & 75.8 & 72.4 \\
\hline
SFT Only & \underline{86.1} & \underline{87.7} & \underline{82.2} \\
RFT Only & 83.7 & 84.2 & 80.2 \\
SFT + RFT & \textbf{93.8} & \textbf{94.6} & \textbf{92.3} \\
\bottomrule
\end{tabular}
\caption{
Reward modeling accuracy (\%) across different training paradigms.
}
\label{tab:performance_ablation}
\end{table}

\begin{figure}[t!]
\centering
\includegraphics[width=0.9\columnwidth]{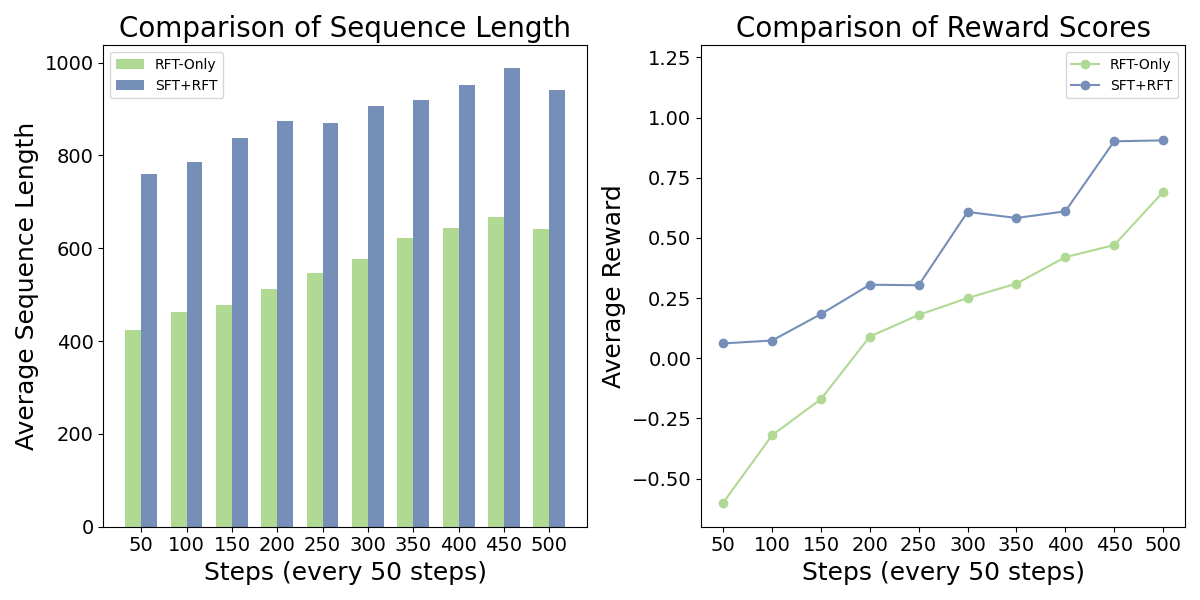}
\caption{Comparison between the RFT stage for the RFT only and SFT followed by RFT training paradigms. 
}
\label{fig:coldrl_vs_perrm}
\end{figure}

To gain deeper insights into the value of the SFT stage, we track the average sequence length and reward scores during the RFT stage for the RFT Only and SFT+RFT training paradigms, as shown in Fig.~\ref{fig:coldrl_vs_perrm}. 
The results demonstrate that performing SFT prior to RFT yields a model capable of generating longer sequences and achieving higher average reward scores compared to the base model, Qwen2.5-7B-Instruct. 
RFT applied to this improved starting point consistently outperforms RFT applied directly to the base model in both sequence length and average reward. 
This underscores the importance of an initial SFT phase for establishing a foundational reasoning capabilities, without which the model struggles to learn effectively.
For more in-depth insights into the RFT process, we provide the training curves for PersRM-R1-3B and PersRM-R1-7B in Appendix.~\ref{app:rft-training-curves}.

\subsection{Cognitive Behaviors Emerge}
We identify all four cognitive behaviors defined in \citet{gandhi2025CognitiveBehaviorsThat} emerging during RFT.
These behaviors impact the reasoning flow, and ultimately contribute to accurate reward score predictions. 
In contrastive, with SFT alone, non of these behaviors except for subgoal setting are observable.
This suggests that the base LLM, i.e., Qwen2.5-7B-Instruct, does not inherently demonstrate these human-like reasoning skills when analyzing personality traits. Instead, these abilities are explored and reinforced during RFT with our constructed preference data, a process analogous to the emergence of the ``Aha moment'' observed in the RFT of DeepSeek-R1 for solving math and coding problems \citep{deepseek-ai2025DeepSeekR1IncentivizingReasoning}. 
Examples of such behaviors are provided in the text box below.

\begin{tcolorbox}[colback=gray!10!white,colframe=blue!65!black,title=Examples of Identified Cognitive Behaviors]
  \textbf{Verification:} ``Let me double-check whether Response A's tone truly aligns with the user's style ...'' \\[4pt]
  \textbf{Backtracking:} ``Initially, I thought Response B matched better, but upon reconsideration, that route doesn't hold due to its formal tone ...'' \\[4pt]
  \textbf{Subgoal Setting:} ``First, let's evaluate tone consistency, then move on to language fluency and finally check personal relevance ...'' \\[4pt]
  \textbf{Backward Chaining:} ``If Response A is indeed superior, it must fulfill all the criteria—let me trace back through each evaluation point to confirm ...''
\end{tcolorbox}

\subsection{Task-specific Behaviors Emerge}

Besides general cognitive behaviors that influence the reasoning flow in general, task-specific behaviors that benefit personality traits analysis also emerge during RFT driven by the optimization objective of accumulative reward maximization. 
Identified emergent abilities include 1) \emph{the discovery of novel and case-specific stylistic criteria} that are not explicitly present in the SFT dataset; and 2) \emph{dynamic and context-aware prioritization for various criteria}.

\noindent\textbf{Discovery of Novel and Case-specific Criteria.} A key capability unlocked by the RFT phase is the discovery of evaluation criteria that are more granular and insightful than the general rubrics provided during SFT. The SFT-only model is confined to applying these pre-defined rubrics from its training data. In contrast, the RL-trained model demonstrates an exploratory capacity, identifying the latent stylistic principles that make an author's voice unique.
For instance, when evaluating responses to the prompt, \emph{``Do you feel that gender discrimination is still an issue in the present-day United States?''}, the reference author constructs their entire argument around a detailed personal anecdote (working as a male secretary).  
The SFT-only model is limited to applying the predefined and constrained evaluation criteria derived from the SFT dataset. However, our SFT+RFT model discovered a more profound emergent criteria: \emph{``Argumentation Grounded in Personal Narrative.''} It learned that the author's core stylistic signature isn't just mentioning an experience, but using it as the foundational structure for the entire essay. Therefore, it correctly identifies that a response that mimics this narrative-driven argumentative structure is far more aligned with the author's style than one that merely presents a generic, albeit well-reasoned, essay on the same topic.

\noindent\textbf{Dynamic and Context-aware Rubric Prioritization.} Unlike the SFT-only model, which applies evaluation criteria in a fixed, predetermined order, the SFT+RFT model dynamically adjusts the priority of evaluation rubrics during inference. It takes into account the specific stylistic context of the reference and the actual content of the responses, allowing it to reason flexibly and emphasize the most relevant criteria for each case. This dynamic prioritization leads to more context-sensitive and accurate preference judgments, rather than relying on a rigid, one-size-fits-all rubric sequence.
For example, its ``<eval>'' block might state: \emph{``Response A is decisively superior as it mirrors the reference author's core technique of `Argumentation Grounded in Personal Narrative.' The entire essay is built around a central story, which is the defining characteristic of the target style. Response B, while addressing the topic, adopts an abstract and impersonal tone, completely missing this crucial stylistic signature.''}
In this example, a specific criterion is deemed dominant and is prioritized in the analysis of personality traits.

These emergent reasoning behaviors during RFT result in both more accurate similarity measure with regard to personality traits and enhanced interpretability. A case study of the reasoning traces generated by PersRM-R1 is presented in Appendix~\ref{app:case-study}.


\section{Discussion}

The superior capabilities of PersRM-R1 in personality trait analysis and similarity measurement open up several promising avenues for future exploration. 
For instance, integrating retrieval-augmented generation (RAG) techniques for exemplar selection could be investigated to further leverage the strengths of PersRM-R1 in scenarios where a set of user demonstrations is available. 
Currently, the training and evaluation of PersRM-R1 are conducted using datasets focused on writing styles, due to the limited availability of open-source user-specific data in other domains. However, the data construction pipeline and training paradigm for PersRM-R1 are not confined to this area and can be an efficiently applied to other domains as well. Scaling and evaluating the approach cross additional domains remains an important direction for future research.


\section{Conclusion}

In this work, we introduce PersRM-R1, the first reasoning-based reward model specifically designed for personalization alignment. 
PersRM-R1 is developed via a novel pipeline that incorporates guided data augmentation, as well as supervised and reinforcement fine-tuning, which internalizes the capability of fine-grained, personality-centric reasoning.
Our experimental results demonstrate that PersRM-R1 achieves state-of-the-art performance in personalized reward modeling with remarkable parameter efficiency. 
Furthermore, PersRM-R1 generalizes robustly across unseen domains and new user exemplars, capturing transferable principles of personal preference.
This work establishes reasoning-based personality analysis as a promising approach for personalized reward modeling, laying a foundation for more adaptive and data-efficient user-aligned LLMs.



\bibliography{mybib, longbib}
\bibliographystyle{iclr2026_conference}

\clearpage
\appendix

\begin{center}
    \Large {APPENDIX}
\end{center}

\section{Preliminary}
\label{app:preliminary}

\subsection{Reward Modeling} \label{subsec:reward_modeling}
Canonical reward models are typically formulated as a Bradley-Terry (BT) models \citep{knox2013LearningNonmyopicallyHumangenerated, christiano2017DeepReinforcementLearning} and are trained in supervised learning.
The model estimates the probability that one response is preferred over another, and the optimization objective is formulated with loss function:
\[
\mathcal{L}_{\text{BT}}(\theta) = - \mathbb{E}_{(x, y^+, y^-) \sim \mathcal{D}} \left[ \log \sigma \left( r_\theta(x, y^+) - r_\theta(x, y^-) \right) \right],
\]
where $\mathcal{D}$ represents a pairwise preference database containing triplet entries $(x, y^+, y^-)$, $x$ is the query, $y^+$ and $y^-$ are the preferred or rejected responses, respectively.
\( r_\theta(x, y) \) denotes the predicted reward score for response \( y \) given input \( x \) by a reward model that is parameterized by $\theta$. 
\( \sigma (\cdot) \) is the sigmoid function.

As causal LLMs exhibit great generalizability across diverse domains \citep{radford2019LanguageModelsAre, brown2020LanguageModelsAre}, reward modeling has been formulated as a generative task recently \citep{liu2025InferenceTimeScalingGeneralist,guo2025RewardReasoningModel}, where a causal LLM works as the reward model to predict the reward score in its generated output. The training objective is identical to the next-token prediction in pre-training:
\[
\mathcal{L}_{\text{Causal}}(\theta) = - \mathbb{E}_{x^\prime \sim \mathcal{D}} \left[ \sum_{t=1}^{T-1} \log p_\theta(x_{t+1} \mid x_1, \ldots, x_t) \right],
\]
where $x^\prime$ is a piece of text constructed with $(x, y^+, y^-)$, $p_\theta(x_{t+1} \mid x_1, \ldots, x_t)$ represents the predicted probability of the next token given previous ones. 
The training sample $x^\prime$ used in this formulation is flexible where intermediate reasoning process can be included before the production of the final reward score.  
We adopt the causal formulation to perform supervised fine-tuning (SFT) in this paper, as it aligns with our goal to incorporate reasoning process into reward modeling. 

\subsection{Reinforcement Fine-tuning}
Proximal policy optimization (PPO) \citep{schulman2017ProximalPolicyOptimization} and its variant group relative policy optimization (GRPO)  \citep{shao2024DeepSeekMathPushingLimits} are two mainstreaming RL methods for RFT to enhance reasoning capabilities of LLMs. 
In this work, we utilize GRPO for RFT following existing work in enhancing reasoning capabilities for LLMs \citep{deepseek-ai2025DeepSeekR1IncentivizingReasoning, chen2025RMR1RewardModeling, guo2025RewardReasoningModel}. 

GRPO is an improved version of PPO designed to boost memory and computational efficiency
Traditionally, PPO uses a value model to estimate the state value of generated responses for advantage estimation \citep{schulman2015HighDimensionalContinuousControl}.
In the domain of RFT for LLMs, this value model is typically initialized with a pretrained LLM similar in size to the policy model and is optimized along with it in supervised learning. 
In contrast, GRPO removes the need for a value model by estimating advantages through a Monte Carlo approach. It calculates advantages based on the rewards from a set of randomly sampled outputs, which greatly reduces both memory and computational cost while ensuring effective policy optimization.

In GRPO, a group of outputs for a given prompt $x$ is generated by sampling from the policy $\pi_\theta$:
\begin{equation}
\{ o_1, o_2, \dots, o_G \} \sim \pi_\theta(\cdot \mid x),
\label{eq:sampling}
\end{equation}
where, $G$ is a hyperparameter that specifies how many outputs are sampled per prompt $x$. 
To estimate the advantage for each sampled output, GRPO uses the normalized reward within the group:
\begin{equation}
A_i = \frac{r_i - \bar{\textbf{r}}}{\sigma(\textbf{r})}, i \in \{1,2,\dots,G\},
\label{eq:advantage-estimate}
\end{equation}
where $\textbf{r} = \{r_1, r_2, \dots, r_G \}$ are the rewards assigned to each output in the group, calculated by a rule-based reward function or a learned reward model. $\bar{\textbf{r}}$ and $\sigma(\textbf{r})$ denote the mean and standard deviation of the rewards in the group, respectively. 
With these advantage estimates, GRPO updates the policy by maximizing the following objective, closely following the PPO framework:
\begin{align*}
\mathcal{J}_\text{GRPO}(\theta) = & \frac{1}{G}\sum_{i=1}^G \frac{1}{|o_i|}\sum_{t=1}^{|o_i|} \biggl\{ \text{min} \Bigl[ \lambda_{t}(\theta) A_{i,t}, \text{clip} \Bigl( \lambda_t(\theta), \\
& 1-\varepsilon, 1+\varepsilon \Bigr) A_{i,t} \Bigr] - \beta \mathbb{D}_{KL}[ \pi_\theta \| \pi_\text{ref} ]   \biggr\},
\end{align*}
where $ \lambda_t(\theta) = \frac{\pi_\theta(o_{i,t} | x, o_{i,<t})}{\pi_{\theta_\text{old}}(o_{i,t} | x, o_{i,<t})}$ measures the change in predicted probability of token $o_{i,t}$ between the current and previous policy models. 
The advantage $A_{i,t}$ is set to $ = A_{i}$ for all tokens in output $o_i$, and $|o_i|$ denotes the number of tokens in output $o_i$.
The hyperparameter $\varepsilon$ controls the clipping range for stable updates, while $\beta$ determines the weight of the KL-divergence constraint with respect to a reference policy $\pi_\text{ref}$. 
For a deeper dive into the theoretical foundations and optimization strategies, we refer readers to the original PPO and GRPO papers \citep{schulman2017ProximalPolicyOptimization, shao2024DeepSeekMathPushingLimits}.

\section{Additional Training and Evaluation Details} \label{app:training_and_testing}
\paragraph{SFT.}
SFT Stage is carried out using the SFTTrainer module from OpenRLHF \citep{hu2025OpenRLHFEasytouseScalable}, with a batch size of 64 and a single training epoch. To improve GPU memory efficiency, we enable gradient checkpointing, leverage FlashAttention, and apply optimizer offloading. The model is optimized using the Adam optimizer with a learning rate of 5e-6. 

\paragraph{RFT.}
We adopt the verl framework \citep{sheng2025HybridFlowFlexibleEfficient} for all GRPO training. The training is performed using Fully Sharded Data Parallel (FSDP) with parameter, gradient, and optimizer offloading enabled to improve memory efficiency. The training batch size is set to 32. Gradient checkpointing is enabled to further reduce memory usage. Rollout generation is handled by the vLLM backend, configured with tensor parallelism size of 1 and GPU memory utilization capped at 0.6. Sampling uses default parameters, with temperature = 1.0 and top-p = 1.0. Each prompt is decoded using 8 candidate responses. KL regularization is applied with a coefficient of 1e-3 and uses the low-variance KL approximation. The learning rate is set to 3e-7 for all model sizes, with no scheduler applied.
We train the 3B and 7B models on 1 node equipped with 8 A100 (80GB) GPUs.

\paragraph{Reward Model Evaluation.} 
The evaluation methodology for RMs is fundamentally dependent on their architecture and task formulation (cf.~Sec.~\ref{subsec:reward_modeling}), which dictates the format of the input data and the method to extract reward scores to calculate reward modeling accuracy. We define distinct input structures for scalar-based and generative models. 

\begin{itemize}
    \item \textbf{Scalar RMs} are trained to predict a quality score for a single response in isolation. The training instances for these models are structured as a \texttt{\{query, response\}} entity, where the model learns to output a scalar value corresponding to an absolute quality rating. To evaluate on pairwise data \texttt{\{query, response\_a, response\_b\}}, the RM performs inference separately on each response in the triplet to obtain individual reward scores. 

    \item \textbf{Generative RMs} are trained on pairwise preference data to directly learn a comparative function. The input for these models is a triplet formatted as \texttt{\{query, chosen\_response, rejected\_response\}}. During training, the model is optimized to assign a higher score to the \texttt{chosen\_response} over the \texttt{rejected\_response} for the same query. For evaluation on pairwise data \texttt{\{query, response\_a, response\_b\}}, the RM processes an input constructed with the triplet and predicts reward scores for both responses simultaneously. 
\end{itemize}

\section{Prompt Examples for Response Sample Generation}
\label{app:promt_examples_sample}

\begin{promptbox}{Prompt Example 1 : Style Mimicking}
You are an expert writer trained to imitate human-written responses. 

Your task is to write a continuation based on the sentence inside <Problem>, while closely following the tone, structure, style, and content of the example inside <Context>.
The example provided does not need to be copied. 

Instead, you should carefully mimic its language patterns, coherence, detail level, and writing flow to generate a similar-quality output.

<Problem>
\textcolor{red}{\{problem\}}
</Problem>

<Context>
\textcolor{red}{\{context\}}
</Context>

Now, please generate a continuation that matches the writing style and quality of the example above, based on the problem.

Your response should read as if it could have been written by the same author who wrote the example in <Context>.
\end{promptbox}

\begin{promptbox}{Prompt Example 2: Minor Replacement}
You must perform only minimal rewriting of the paragraph below. Strictly replace exactly 5 to 6 individual words only, preferably adjectives, adverbs, or verbs, with close synonyms. 

Do not change sentence structure, punctuation, or the order of words. 

Do not alter nouns, names, numbers, or proper terms.  

Do not insert or remove any words — the total word count should remain nearly identical. 

The meaning, tone, and style of the paragraph must remain unchanged.

Do not explain your changes — only output the rewritten paragraph.

Here is the paragraph to rewrite:
\textcolor{red}{\{paragraph\}}
\end{promptbox}

\begin{promptbox}{Prompt Example 3 : Random Style}
You are a creative writer. 

Your task is to randomly continue the given sentence in any style, tone, or direction. 

Each time you see the same input, you must write something completely different from before. 

You are encouraged to be unpredictable, humorous, absurd, poetic, or even surreal.

Avoid logical consistency or factual accuracy. 
Focus on creativity, stylistic variation, and divergence from expected content.

The sentence you must continue is shown below between <Problem> tags. 
Your response should be a free-form continuation, based solely on the content within <Problem>.

<Problem>
\textcolor{red}{\{problem\}}
</Problem>

Now, please generate the continuation.

\end{promptbox}

\section{Prompt Examples for Reasoning Trace Generation} \label{app:promt_examples_reason}
\begin{promptbox}{Prompt Example 4: Reasoning Trace Generation}
You are a skilled assistant for scoring and comparing responses. 

You need to evaluate two given responses based on how well they match the context's personal style, tone, preferences, and writing style.

Your overall scores must reflect a single, comprehensive judgment per response that integrates all criteria.

Personal-Align-Specific Criteria (you may choose from or add to these as appropriate):

1. Personal Style Adherence:

2. Tone and Voice Consistency:

3. Language Fluency and Coherence:

4. Relevance to Personal Preferences and Experiences:

You may also add other relevant criteria if they help evaluate the two responses.

Your task is to use the most suitable criteria for the current context to judge the responses holistically and then output one overall integer score per response.

Conversation Context: \textcolor{red}{\{ context \}}

Responses to be Scored: \\

Response A: \textcolor{red}{ \{ response a\} }

Response B: \textcolor{red}{ \{ response b\} }

Output Format Requirements
Produce your answer as exactly three sections:

<criteria>  Specific Criteria: Clearly list the criteria you decided to use for this context (you can select from the ones provided or add new ones if appropriate). Do not include any scores in this section.</criteria>

<eval>  Analysis: Provide a very detailed, thorough, and step-by-step reasoning process. Break your reasoning into multiple points as you go through each chosen criterion one by one. Explain how Response A and Response B satisfy or fail that criterion, and clearly state the impact of each observation on your final judgment. End this section with a concluding sentence that ties all observations together and justifies the overall scores you will give. Do not include any numeric scores in this section.</eval>

<scores>  
Scores: [[x,y]]</scores>
\end{promptbox}

\section{Case Study} \label{app:case-study}

In this section, we present a concrete example generated by PersRM-R1, accompanied by explanatory comments (highlighted in gray) to clarify its behavior. These annotations are added for illustration purposes and are not part of the original model output.

\begin{promptbox}{Response Generated by PersRM-R1}
\textbf{<criteria>} \\
Specific Criteria:
\begin{enumerate}
    \item Personal Style Adherence
    \item Tone and Voice Consistency
    \item Language Fluency and Coherence
    \item \textcolor{red}{Argumentation Grounded in Personal Narrative} \textcolor{gray}{\# Note: PersRM-R1 discovers novel sample-specific criteria that are beyond those included in synthetic data generation for SFT. }
\end{enumerate}
\textbf{</criteria>}

\bigskip
\hrule
\bigskip

\textbf{<eval>} \\
\textbf{Analysis:}

\textcolor[HTML]{2962FF}{First, let's establish a clear evaluation plan: I will assess the four standard criteria sequentially, then focus on the more nuanced, discovered criterion, and finally, trace back through all points to confirm the final scores.} \textcolor{gray}{\# Note: Cognitive Behavior: Subgoal Setting}

\begin{enumerate}
    \item \textbf{Personal Style Adherence:}
    \begin{itemize}
        \item \textbf{Response A:} This response closely mirrors the user’s personal style. It effectively integrates casual language and humor, making it feel like a natural continuation of the user’s thoughts.
        \item \textbf{Response B:} This response also adheres well, but is slightly more formal in some parts, representing a minor deviation from the user’s typically more relaxed style.
    \end{itemize}

    \item \textbf{Tone and Voice Consistency:}
    \begin{itemize}
        \item \textbf{Response A:} The tone is fully aligned. It maintains the same level of emotional intimacy and casualness as the user’s original message, feeling natural and consistent.
        \item \textbf{Response B:} The tone is mostly aligned. However, the slight formality can make it feel a bit less personal and more distant compared to the user's voice.
    \end{itemize}

    \item \textbf{Language Fluency and Coherence:}
    \begin{itemize}
        \item \textbf{Response A:} The language is very fluent and smooth, with no awkward phrasing.
        \item \textbf{Response B:} Mostly fluent, but the intermittent formality creates a slightly disjointed reading experience compared to the user's seamless style.
    \end{itemize}

    \item \textbf{Argumentation Grounded in Personal Narrative:} This is the crucial differentiator that goes beyond surface-level style matching. It's not just about *including* anecdotes (covered by 'Relevance'), but about using a personal story as the foundational structure for the entire argument. This insight represents the \textcolor[HTML]{6A1B9A}{Discovery of Emergent and Naunced Criteria}. In this context, this criterion is weighted heavily, demonstrating \textcolor[HTML]{6A1B9A}{Dynamic and Context-Aware Rubric Prioritization}.
    \begin{itemize}
        \item \textbf{Response A:} Excels here. It correctly identifies and replicates the user’s method of building the entire response around a central personal story.
        \item \textcolor[HTML]{2962FF}{Initially, I thought Response B’s use of examples was sufficient. However, upon reconsideration, that route doesn't hold. Response B only sprinkles in anecdotes as supporting evidence for a list of points, failing to grasp that the narrative *is* the argument. This is a fundamental stylistic misunderstanding.} \textcolor{gray}{\# Note: Cognitive Behavior: Backtracking}
    \end{itemize}
\end{enumerate}

\textbf{Concluding Summary:}
\textcolor[HTML]{2962FF}{Let me double-check the overall logic before scoring.} (Cognitive Behavior: Verification) While both responses perform adequately on surface-level criteria like tone and fluency, Response A demonstrates a profoundly deeper understanding of the author's style by correctly identifying and replicating the core \textbf{Argumentation Grounded in Personal Narrative} technique. Response B's failure on this single, prioritized criterion creates a significant quality gap. \textcolor[HTML]{2962FF}{Therefore, if Response A is superior, it must have fulfilled this key requirement, which it has. Tracing back, this single point justifies a much higher score.} (Cognitive Behavior: Backward Chaining) This clear, verifiable distinction based on a critical discovered criterion perfectly exemplifies \textcolor[HTML]{6A1B9A}{Enhanced Interpretability and Faithful Adherence} and is the basis for the final scores.
\textbf{</eval>}

\bigskip

\textbf{<scores>}
[[9,7]]
\textbf{</scores>}

\end{promptbox}

\section{Additional Experimental Results}
\subsection{Evaluation of Constructed Preference Dataset}
In this part, we evaluate the quality of the preference dataset constructed after the first stage of our data curation pipeline (see Fig.~\ref{fig:persrm-r1-data generation pipeline}, left panel). 
Following previous work \citep{zheng2023JudgingLLMasaJudgeMTBench, shaikh2025ShowDontTell, yang2024RLCDReinforcementLearning}, we use GPT-4o as a proxy for human judges to compare the stylish similarity between responses and personal exemplars. 
GPT-4o is prompted to produce a similarity score given a pair of responses over 300 randomly selected samples for each of the 5 methods, for a total of 1,500 samples as our evaluation metric, where a higher similarity score indicates closer personality traits exhibited in the given pair of responses. The prompt used for this evaluation is provided below. Specifically, GPT-4o is prompted to assign a similarity score to each response and exemplars pair on 1,500 samples. A higher similarity score reflects a greater alignment in the exhibited personality traits between the response and the personal exemplars. The prompt used for this evaluation is presented below.
\begin{promptbox}{Prompt Example for Style Similarity Evaluation }
 Evaluate the similarity between the following two sentences in terms of sentence structure, grammar, style, tone, and word choice. 

Give a comprehensive similarity score ranging from 0 to 10, where a higher score indicates greater similarity. 

Only return a single number. Do not provide any explanation.
\end{promptbox}
Table~\ref{tab:style_similarity_scores} shows that different generation strategies yield clearly distinguishable similarity scores, confirming the effectiveness and stylistic separability of our preference data. The high scores for target-author writings and low scores for randomized styles suggest that the dataset can provide reliable training signals for style alignment.
\begin{table}[h]
\centering
\begin{tabular}{|l|c|}
\hline
\textbf{Category} & \textbf{GPT-4o-eval score} \\
\hline
Other writings by the target author & 9.41 \\
Minor Replacement & 9.39 \\
Style Mimicking & 5.89 \\
Writings by other authors & 3.86 \\
Style Randomization & 2.41 \\
\hline
\end{tabular}
\caption{Style similarity scores evaluated by GPT-4o}
\label{tab:style_similarity_scores}
\end{table}

\subsection{RFT Training Curves} \label{app:rft-training-curves}
\begin{figure}[h]
\centering
\includegraphics[width=0.65\columnwidth]{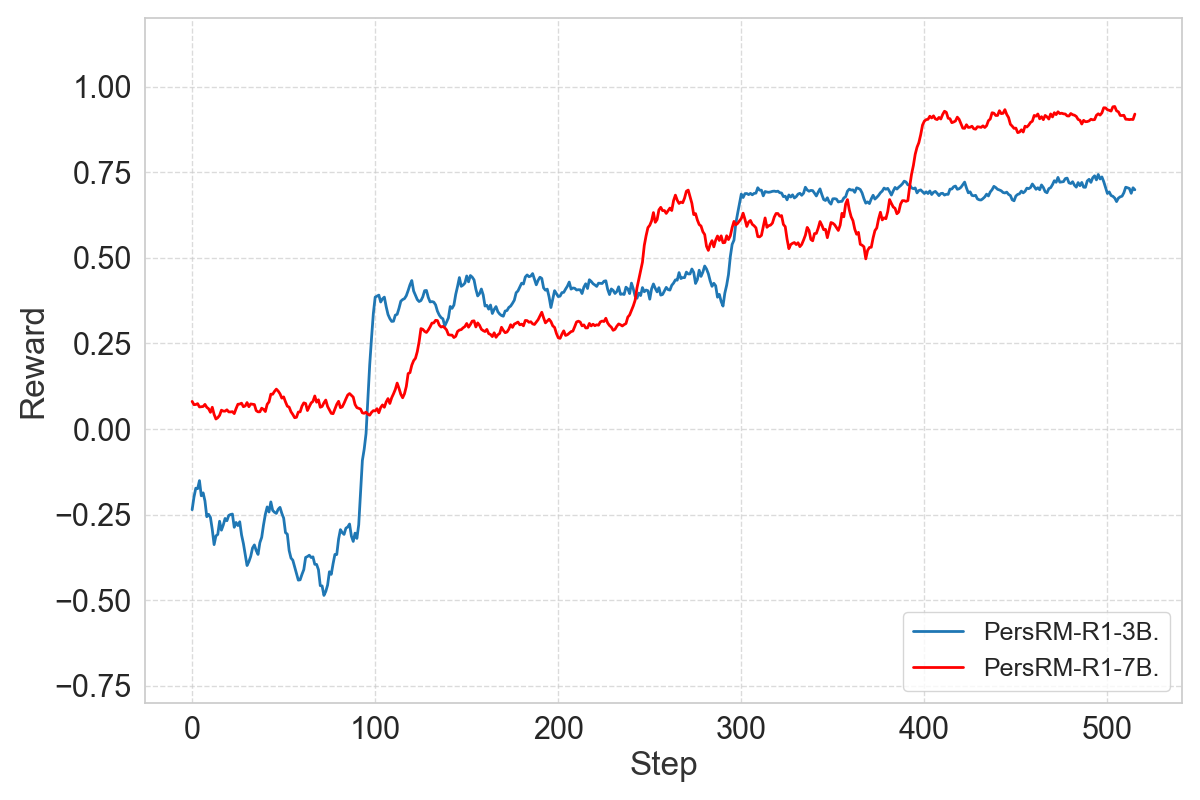}
\caption{Reward curves during RFT for PersRM-R1-3B and PersRM-R1-7B.}
\label{fig:training-curve}
\end{figure}

Fig.~\ref{fig:training-curve} illustrates training curves of RFT during the developing of PersRM-R1-3B and PersRM-R1-7B. 
The average reward obtained by each model (cf.~Eq.~\ref{eq:reward_function}) steadily increases until reaching a plateau.
Notably, PersRM-R1-7B achieves a higher average reward than PersRM-R1-3B at both the initial stage and at convergence, indicating its superior final performance after both SFT and RFT. 
Interestingly, the smaller model surpasses the larger model in terms of reward during steps [100, 250] and [300, 400], though the larger model eventually overtakes it. 
These curves clearly illustrate the effectiveness of our RFT training in enhancing the model's capability to analyze personality traits and produce reward scores that reflect personality similarity. 

\subsection{Generalizability to Additional Exemplars of Different Training Paradigms} \label{app:three-shot-generalizability-training-paradigms}
We evaluate the generalizability to additional exemplars in inference time for models that are trained in different training paradigms (cf.~Sec.~\ref{sec:training-paradigms}) using Qwen2.5-7B-Instruct as the base model. Experimental results are presented in Table~\ref{tab:three-shot-training-paradigms}. 
This demonstrates that the model trained through SFT followed by RFT, i.e., PersRM-R1-7B, exhibits stronger generalizability to additional exemplars.

\begin{table}[ht]
\centering
\footnotesize
\begin{tabular}{cllll}
\toprule
\textbf{Training Paradigm} & \textbf{\# Ex.} & \textbf{CCAT} & \textbf{CMCC} & \textbf{Cross-Domain} \\
\midrule
\multirow{2}{*}{SFT Only} & 1 & 86.1 & 87.7 & 82.2 \\
 & 3 & \underline{86.3} {(+0.2)} & \underline{87.7} (+0) & \underline{82.4} (+0.2) \\
\hline
\multirow{2}{*}{RFT Only} & 1 & 83.7 & 84.2 & 80.2 \\
 & 3 & 83.7 (+0) & 84.2 (+0) & 80.5 (+0.3) \\
\hline
\multirow{2}{*}{SFT + RFT} & 1 & {93.8} & {94.6} & {92.3} \\
& 3 & \textbf{94.1} {(+0.2)} & \textbf{95.1} {(+0.5)} & \textbf{92.6} {(+0.4)} \\
\bottomrule
\end{tabular}
\caption{Reward modeling accuracy (\%) with varying number of exemplars (\# Ex.).  
Values in parentheses indicate the improvement achieved relative to using one exemplar.
The best-performing results in the three-exemplar setup are highlighted in \textbf{bold}, while the runner-up results are \underline{underlined}.
}
\label{tab:three-shot-training-paradigms}
\end{table}

\section{Dataset Details} \label{app:dataset}

Our dataset is constructed from two corpora: \textbf{CCAT} (news) and \textbf{CMCC} (multi-genre). We adopt a strictly author-disjoint split to ensure that the evaluation measures the model's generalizability. Details of dataset splits are provided in Table~\ref{tab:dataset_splits}.

\begin{table}[ht]
\centering
\footnotesize
\begin{tabular}{lccc}
\toprule
\multirow{2}{*}{\textbf{Split}} & \textbf{\# Authors} & {\textbf{\# Pairwise}} \\
& \textbf{(CCAT / CMCC)} & \textbf{Samples} \\
\midrule
Training Set & 45 / 18  & 17.2k \\
Validation Set & 2 / 1  & 200 \\
Test Set (In-domain) & 3 / 2  & 334 \\
Test Set (Cross-Domain) & 0 / 3 & 439 \\
\bottomrule
\end{tabular}
\caption{Author and sample counts of each dataset split. }
\label{tab:dataset_splits}
\end{table}

\end{document}